\documentclass[letterpaper, 10 pt, conference]{ieeeconf}  % Comment this line out if you need a4paper

\IEEEoverridecommandlockouts                              % This command is only needed if 
\usepackage{times}
\usepackage{graphicx}
\usepackage{multirow}
\usepackage{amsmath}
\usepackage{subcaption}
\usepackage{comment}
\usepackage{color}
\title{\LARGE \bf
Semantics-STGCNN: A Semantics-guided Spatial-Temporal Graph Convolutional Network for Multi-class Trajectory Prediction
}

\author{Ben A. Rainbow$^{1}$ Qianhui Men$^{2}$ and Hubert P. H. Shum$^{1*}$% <-this % stops a space
\thanks{$^{1}$B. A. Rainbow and H. P. H. Shum are with the Department of Computer Science, Durham University, UK.
        {\tt\small \{ben.a.rainbow, hubert.shum\}@durham.ac.uk}}%
\thanks{$^{2}$Qianhui Men is with the Department of Computer Science, City University of Hong Kong, Hong Kong.
        {\tt\small qianhumen2-c@my.cityu.edu.hk}}%
\thanks{*Corresponding author}% <-this % stops a space
}

\begin{document}

\maketitle
\thispagestyle{empty}
\pagestyle{empty}

%%%%%%%%%%%%%%%%%%%%%%%%%%%%%%%%%%%%%%%%%%%%%%%%%%%%%%%%%%%%%%%%%%%%%%%%%%%%%%%%
\begin{abstract}
Predicting the movement trajectories of multiple classes of road users in real-world scenarios is a challenging task %with diverse trajectory patterns 
%traffic 
due to the diverse trajectory patterns. 
While recent works of pedestrian trajectory prediction successfully modelled the influence of surrounding neighbours based on the relative distances, they are ineffective on multi-class trajectory prediction. 
This is because they ignore  
%A successful prediction model requires to produce accurate trajectories across a number of classes of objects.
% problem of recent work
%However, recent work only modelled the importance of surrounding objects using metrics based on the fixed distances, 
%while ignoring 
the impact of the implicit correlations between different types of road users on the trajectory to be predicted---for example, a nearby pedestrian has a different level of influence from a nearby car.
In this paper, we propose to introduce class information into a graph convolutional neural network to better predict the trajectory of an individual. We embed the class labels of the surrounding objects into the label adjacency matrix (LAM), which is combined with the velocity-based adjacency matrix (VAM) comprised of the objects' velocity, thereby generating a semantics-guided graph adjacency (SAM). SAM effectively models semantic information with trainable parameters to automatically learn the embedded label features that will contribute to the fixed velocity-based trajectory. Such information of spatial and temporal dependencies is passed to a graph convolutional and temporal convolutional network to estimate the predicted trajectory distributions. We further propose new metrics, known as Average\textsuperscript{2} Displacement Error (aADE) and Average Final Displacement Error (aFDE), that assess network accuracy more accurately. We call our framework Semantics-STGCNN. It consistently shows superior performance to the state-of-the-arts in existing and the newly proposed metrics.
\end{abstract}


\section{Introduction}
Trajectory prediction is attracting increasing attention. % with the development of autonomous vehicles. 
Accurate trajectory prediction in autonomous driving \cite{autocar} allows the car to plan a better trajectory for itself as it can accurately predict where other objects in the scene plan to go. It can also help surveillance systems \cite{surveillance} to identify suspicious activity or dangerous situation in the traffic and alert in advance. 
%With the emergence of COVID-19 many countries have introduced social distancing laws with many restricting the number of people that can move as a group. Many approaches such as \cite{socialrecursivebehaviour} can identify individuals moving as groups. This could allow law enforcement officers to find groups breaking such laws to help slow the spread of COVID-19.

However, trajectory prediction is challenging, especially for dealing with multiple classes of road users, such as \textit{pedestrians}, \textit{cars}, and \textit{bikers}. Existing works, such as Social-LSTM~\cite{mohamed2020social} that learn temporal trajectory dependencies with recurrent architecture, or Social-GAN~\cite{gupta2018social} that increases the diversity of trajectory with the generative adversarial network (GAN)~\cite{goodfellow2014generative}, model social correlations for pedestrians only. Such models are less extensible to scenarios with multiple classes. Another problem raised is how to accommodate different classes of trajectories so that they can learn from each other for a better prediction. This is based on the observation that different object classes have different influences on the predicting trajectory. For example, even with the same velocity, a car coming close to a pedestrian has a different influence on a bicycle. State-of-the-art models such as Social-STGCNN~\cite{mohamed2020social} are ineffective in such scenarios as they only look at the trajectory features without identifying the types of trajectories.

In this paper, we propose a new framework called Semantics-STGCNN, which is a spatial-temporal graph convolutional network for effective multi-class trajectory prediction. %by aim to enhance the predicted trajectory with more informative social intersections 
%by introducing the trainable class label embeddings with the prediction in a supervised manner. 
Our main insight is to embed class labels into the adjacency matrix of the graph convolutional network, such that object class information can be harnessed to inform the prediction.
Such an insight is implemented through  
%a Semantics-guided Spatial-Temporal Graph Convolutional Network (Semantics-STGCNN) for predicting trajectories under different classes. By regarding each object in a scene as a graph node, the core of our model is 
a semantics-guided graph adjacency matrix (SAM), which represents the re-weighted social correlations from the high-level semantic meanings of object types. SAM is generated by combining a label-level adjacency matrix (LAM) that functions like a trainable attention matrix based on the class labels with a non-trainable velocity-based graph adjacency matrix (VAM) under the fixed trajectory. This enables the latter to automatically tell the important correlations in the represented graph based on the nature of the object class. %More specifically, the LAM and SAM are flexibly trained with the features integrated from fully-connected layers.

Together with the learned graph structure of SAM, the trajectory features are fed into a spatial-temporal graph convolutional network (ST-GCNN) to model the inner dependencies among the trajectory nodes, followed by a temporal extrapolator convolutional neural network (TXP-CNN) to capture the in-depth temporal dependencies and to estimate the bi-variant distributions for every future time step. The steps of ST-GCNN and TXP-CNN are inspired by Social-STGCNN~\cite{mohamed2020social}, which designs a well-organized model to explore effective spatial and temporal representations of the social interactions among the trajectories. 

To accurately evaluate the performance of the framework, together with existing metrics, we propose novel metrics to evaluate the predicted distribution known as Average\textsuperscript{2} Displacement Error (aADE) and Average Final Displacement Error (aFDE). They provide a more realistic measurement to justify the distributions by calculating the average error across a number of sampled predictions, compared to the existing metrics that only sample the prediction with the minimum error.

The experiments are conducted quantitatively and qualitatively based on the real-world trajectory dataset with multiple types of trajectories. The quantitative results show that by informing the model with high-level embedded label information, our proposed Semantics-STGCNN produces future trajectories with lower error accumulations under various numerical metrics. The qualitative results further show that the proposed method can generate more accurate future motions for both moving and static objects. 

The contributions of the paper is summarized as follows:
\begin{itemize}
\item We propose Semantics-STGCNN, a semantics-guided spatial-temporal graph convolution network for multi-class trajectory prediction. Core to the network is the construction of an adjacency matrix that is informed by the semantics of the trajectory data based on the embedded class labels.
\item We propose new metrics known as Average\textsuperscript{2} Displacement Error (aADE) and Average Final Displacement Error (aFDE), and demonstrate that this metric provides a more robust evaluation on the accuracy of trajectory prediction networks.
\item We open our source code for validation and further development, which can be found on https://github.com/Yutasq/Multi-Class-Social-STGCNN
\end{itemize}

\section{Related Work}
%The recent increase in work relating to autonomous driving has lead to an increase in pedestrian trajectory prediction solutions. 
With the rapid advancement of deep learning, the model capacity of understanding and predicting a large number of trajectories has increasingly improved. Here, we recall the existing work in trajectory prediction that is highly related to the theme and the method proposed in this paper.
%however the field started with the idea of social forces and handcrafted equations that defined pedestrian movement. 

\subsection{Social Modeling in Trajectory Prediction}
A pioneering work in the field of pedestrian movement is \cite{yamaguchi2011you}, which introduces a pedestrian motion model where pedestrians are exerted upon by ``social forces" as a measure for the internal motivations of the individuals to perform certain actions. This model achieves competitive results on modern pedestrian datasets under relatively large scales. Later, many advanced approaches~\cite{gupta2018social,alahi2016social,sadeghian2019sophie,rudenko2020human} are proposed to detect the social correlations based on \cite{yamaguchi2011you}. One remarkable variation comes from \cite{robicquet2016learning} that introducing ``social sensitivity" to characterize the distance between two interacted targets. Such a model helps to define navigational styles and initiate multi-target tracking. Social-LSTM~\cite{alahi2016social} further generalizes to more complex crowded scenes, where recurrent networks~\cite{hochreiter1997long} are first used to learn the state of the person and predict the future trajectory. The spatial correlations of pedestrians are modelled through a new pooling strategy where the hidden states of LSTMs are shared among its neighbours. To increase the variations in the generated trajectories, Social-GAN~\cite{gupta2018social} introduces variety loss by encouraging the generative network to spread its distribution and cover the space of possible paths. 

\begin{figure*}
  \makebox[\textwidth][c]{\includegraphics[width=1\textwidth]{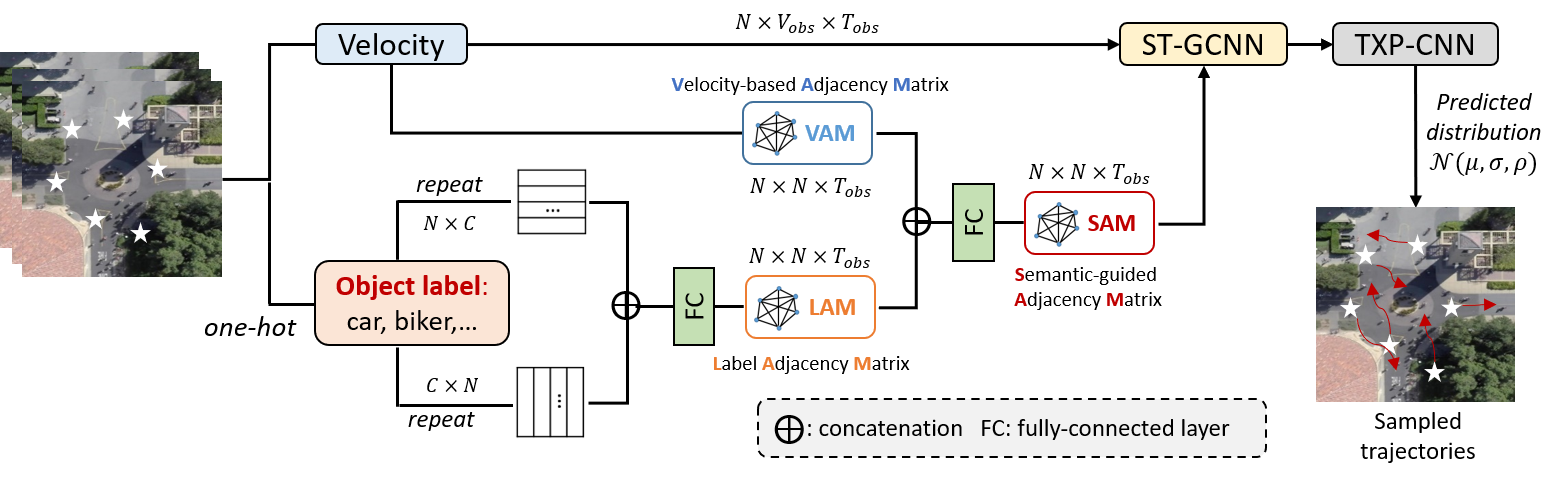}}%
  \caption{The network model for Semantics-STGCNN. For a given sequence of $T_{obs}$ frames with $N$ objects, a velocity-based adjacency matrix (VAM) with the relative coordinates of consecutive frames is created. We then propose a label adjacency matrix (LAM) from the class label of length $C$, combined with VAM to create the final embedding of the semantics-guided adjacency matrix (SAM) for the following graph convolutional neural network. Such an addition informs the upcoming ST-GCNN and TXP-CNN layers with the class information of a trajectory.}
\label{fig:model}
\end{figure*}

\subsection{Graph Neural Networks in Trajectory Prediction}
The previous methods that use aggregation layers based on heuristics like pooling show limited capacity in modeling interactions between pedestrians, and with the emergence of graph neural networks, such problem is addressed by encoding the trajectory in the representative graph structure to handle the social interactions with implicit correlations. For example, a spatial-temporal graph attention network~\cite{stgat} is devised to capture the informative graph representations with both spatial and temporal correlations between pedestrians, and later Social-BiGAT~\cite{kosaraju2019social} extends it to improve predictions in multi-modal situations with GAN. To fully take advantage of the graph representation, the very recent Social-STGCNN~\cite{mohamed2020social} models the scene as a spatial-temporal graph. This replaces the need for aggregation layers as the edges of the graph represent the interactions between two individuals. The graph is then fed into a graph convolutional neural network~\cite{kipf2016semi} to extract the features utilized by a temporal CNN~\cite{lea2017temporal} to predict trajectories. 

%Another piece of work that was developed during the same time looked at a Recursive Social Behaviour Graphs \cite{socialrecursivebehaviour}. This paper outperformed the previous state of art, though it was improved upon by Social STGCNN \cite{mohamed2020social}. This confirms that graph convolution is now the way to achieve the best results for predicting pedestrian trajectories as two Graph Convolutional Neural Networks have outperformed the previous state of the art by at least 10\%. Though this paper may not be as accurate at predicting trajectories as Social STGCNN it added some interesting novel ideas on grouping pedestrians to explore the relationships between them. The paper introduces a neural network to recursively extract social relationships and formulate them into a social behaviour graph called \textit{Recursive Social Behavior Graph for Trajectory Prediction} \cite{socialrecursivebehaviour}. Each pedestrian is considered a node in this graph with features that consider historical trajectories. These nodes are connected by relational social representations which are edges. This is the first time social related annotations have been used to help neural networks learn social relationships. A Graph Convolutional Neural Network is then applied to the graph to extract features which is passed to both a Bicycle LSTM (BiLSTM) to perform trajectory prediction and a Graph Convolutional Neural Network to group individuals. This saw a 10\% improvement over STGAT \cite{stgat}, the previous state of the art, compared to social STGCNN's 20\%.

\subsection{Multi-Class Trajectory Prediction}
So far, a lot of work has gone into pedestrian trajectory prediction, and there have been great leaps and advances in the models used and their subsequent results. The world, however, is a large multi-class system where the trajectory of one target depends on the trajectory of a large number of other objects, all of which can be a variety of classes. Therefore, the models based on a single type of trajectory do not provide a real-world scenario where targets of multiple classes are interacting with each other over large distances. For the models targeting multi-class trajectory prediction either discard the other class data as in \cite{robicquet2016learning,longtermpred} or treats all trajectories as a single class as in \cite{robicquet2016learning,pecnet}. Recently, DESIRE~\cite{lee2017desire} proposes a variational auto-encoder to estimate the optimal policy by looking at the rewards of the predicted trajectory. CAR-Net~\cite{sadeghian2018car} attaches attention modules on the recurrent framework to detect the salient parts of the scene from the spatial contexts to create a more accurate predicted path for various types of trajectories. %Another recent paper introducing PECNet \cite{pecnet} achieves outstanding performance 
However, the models mentioned above essentially ignore important class data about the targets that could help inform the model and improve the predicted trajectories.

\section{Semantics-STGCNN}
The proposed model consists of four components: velocity-based graph learning, semantic-based graph learning, graph convolution, and time extrapolator CNN (TXP-CNN). The velocity-based graph learning module turns a series of past positions into relative trajectories and a similarity matrix between all the objects in the scene. The semantic graph learning module generates a label adjacency matrix (LAM) by encoding a series of class labels in the scene into an appropriate representation and correlates the velocity-based adjacency (VAM) with semantic meanings, the combination of which forms a semantics-guided adjacency matrix (SAM) to enrich the graph structure. The generated semantics-guided graph representation is further used to extract features with the spatial-temporal graph convolution operations (ST-GCNN) and temporal extrapolator CNN (TXP-CNN) following Social-STGCNN~\cite{mohamed2020social} to predict the future trajectories. The output trajectories are represented by a bi-variate Gaussian distribution for which the model outputs a mean $\mu$, standard deviation $\sigma$, and correlation $\rho$ for each object in the scene, which represents the trajectory distribution of each object. Figure \ref{fig:model} visualizes the network model of the proposed framework.
%The graph convolution module is a spatial-temporal graph convolution neural network which conducts spatial-temporal convolution operations on the objects based on the generated semantics-guided graph representation to extract features. These features are a compact representation of the observed trajectory history \cite{mohamed2020social}. Finally the time extrapolator CNN reconstructs and predicts the future trajectories of all objects in the scene. 

\subsection{Problem Formulation}
Given a set of $N$ objects in a scene with their corresponding observed coordinates over a number of time steps $T_{obs}$, our aim is to predict coordinates of each object after $T_{pred}$ time steps. The $n$-th predicted trajectory is denoted as $p_{t}^{n} = (x_{t}^{n}, y_{t}^{n})$ with $t\in \{T_{obs}+1,...,T_{obs}+T_{pred}\}$, where $(x_{t}^{n}, y_{t}^{n})$ are random variables describing the probability distribution of the pixel location of the object $n$ at time $t$ in the 2D image. We assume that $(x_{t}^{n}, y_{t}^{n})$ follows a bi-variate Gaussian distribution $p^n_t \sim N(\mu^n_t, \sigma^n_t, \rho^n_t)$ as in \cite{mohamed2020social}.

\subsection{Velocity-based Graph Representation}
The velocity-based graph representation is defined on the relative positions of the object trajectory to capture the geometric dependencies in the scene, as in \cite{mohamed2020social}. Given a scene with $N$ objects at frame $t$, their trajectories are represented as the graph format $G_t=(V_t, E_t)$ with $t \in \{1,...,T_{obs}\}$, where $|V_{t}|=N$ is the vertex set with $N$ nodes, and $V_t=\{v_t^i\}=\{(x^i_{t}-x^i_{t-1}, y^i_{t}-y^i_{t-1})|\forall i\in\{1,...,N\}\}$ representing the velocity of the consecutive frames. $E_t$ is the set of edges that models the geometric correlations among nodes within the graph $G_t$, and it is reflected by an $N\times N$ adjacency matrix $A^{velo}_{t}$ (i.e. VAM) at each time step:
\begin{equation} 
A_t^{velo}(i,j) =\begin{cases} 
	1/||v^i_t - v^j_t||_2 & \text{if } ||v^i_t - v^j_t||_2 \neq 0, \\
	0 &  Otherwise.
\end{cases}
\label{eq:adj_matrix_sim}
\end{equation}

Here, we use inverse Euclidean distance to define $A^{velo}_t$, as objects that are further away are less likely to have an impact on the trajectory.

\subsection{Semantics-guided Graph Representation}
We then propose a semantics-guided graph representation to enhance the velocity graph with label information to improve the quality of the trajectory interpretation. %This is done by combining the aforementioned velocity-based adjacency matrix (VAM) and an innovative label-oriented adjacency matrix (LAM) such that the final adjacency (SAM) can more closely connect the nodes within the same class. 
This is done by training an embedding of the class label with supervision to inform how much a class of object would affect another class, and represent that kind of relationship in a semantics-guided adjacency matrix (SAM). For example, if a person sees a car running close, the movement will be more affected than when the person sees another person running close. %For example, bikers usually have similar speed with each other compared to the slower pedestrians or the faster cars. 
Modeling label information from different types of objects can greatly influence the relationship between trajectories, which is always ignored by existing works.

\subsubsection{Label Adjacency Matrix (LAM)}
We first propose a label encoding module to derive the label-based graph representation. The objective is to encode the label from supervised embedding such that we inform the VAM which class would affect more to the current trajectory. This is done by creating an adjacency matrix with trainable parameters containing the label information (LAM), the value of which indicating the relationship of the two intersecting object's classes.

More specifically, with the observation that the label embedding is independent of objects, i.e. two objects with the same label should be treated equally, we construct a label-based graph for different objects only from label information. To achieve this goal, we first convert the label of each object to one-hot encoding, where each vector is of length $C$ equal to the number of classes. The one-hot embedding is an effective format of incorporating the discrete features such as labels~\cite{zhang2020semantics}: only one variable is assigned to be $1$ with the position indicating its own class, and the rest are assigned with $0$s. This is done for all $N$ objects in the scene to create an $N\times C$ tensor $L$. We then repeat the one-hot encoding $L$ and its transpose $L^{'}$ for $N$ times at column and row respectively to create an $N\times N$ tensor, where each element consists of $C$ dimensions of the labeling. These two resulted tensors are then reshaped and concatenated in the label dimension to produce the intersection tensor with the size of $N\times N\times 2C$, with an example shown in Table~\ref{tab:onehotenccomb}. We then consider a fully-connected layer to scale down the intersection tensor to a similarity matrix (i.e. LAM) with trainable parameters to automatically encode the useful features from the label information that will contribute to the trajectory relationship. By integrating the pairwise label encoding, the reduced label adjacency matrix (LAM) $A^l$ demonstrates the semantic-level relationship between any pairs of object labels. 

\begin{table}
\centering
\begin{tabular}{c|cccc}
\cline{1-4}
               Object ID                 & $Ped_1$       & $Biker_1$     & $Ped_2$       &  \\ \cline{1-4}
\multicolumn{1}{c|}{$Ped_1$}   & {[}0,1,0,1{]} & {[}1,0,0,1{]} & {[}0,1,0,1{]} &  \\
\multicolumn{1}{c|}{$Biker_1$} & {[}0,1,1,0{]} & {[}1,0,1,0{]} & {[}0,1,1,0{]} &  \\
\multicolumn{1}{c|}{$Ped_2$}   & {[}0,1,0,1{]} & {[}1,0,0,1{]} & {[}0,1,0,1{]} &  \\ \cline{1-4}
\end{tabular}
\caption{Example intersection tensor with 3 objects of two classes \textit{pedestrian} and \textit{biker} representing 3 nodes in the graph.}
\label{tab:onehotenccomb}
\end{table}

\subsubsection{Semantics-guided Adjacency Matrix (SAM)}
With the observation that the trajectory depends on not only the distance to other objects but also the types of objects interacting, we introduce the object label into the graph adjacency matrix $A^{velo}$ for trajectory prediction. This is inspired by \cite{zhang2020semantics} fusing the label of body joint to represent human skeleton graph, which has achieves superior performance for action recognition tasks by incorporating the semantic meanings of the node information.

In particular, we merge the label-oriented correlation $A^l$ and the velocity-based correlation $A^{velo}$ for the final connectivity map of the objects in the scene, which is represented by a semantics-guided adjacency matrix (SAM) $A^s$. To this end, we concatenate $A^{l}$ and $A^{velo}$ in the feature dimension and pass the concatenated adjacency through another fully-connected layer to produce the resulting $A^s$ of size $N\times N\times T_{obs}$. The function of this fully-connected layer is to provide further trainable elements to increase the capacity of the system with better prediction accuracy, which works similar to an attention mechanism that tells the salient correlations within VAM to predict the trajectories.

\subsection{The Spatial-Temporal Graph Convolution Neural Network (ST-GCNNs)}
Spatial-Temporal Graph convolution is then applied to integrate the graph representation of the objects using a spatial-temporal convolution operation as introduced in \cite{mohamed2020social}. The ST-GCNN is defined on graph $G$ at different timestamps, which represents the spatial-temporal information of the scene. With $A^s$ denoting the graph topology at frame $t$, we first generate the laplacian matrix by normalizing $A^s$:
\begin{equation} 
\hat{A}^s_t = \Lambda_t^{-\frac{1}{2}} (A^s_t+I) \Lambda_t^{-\frac{1}{2}},
\label{eq:anorm}
\end{equation}
in which we consider the self-loop of nodes by adding the identity matrix $I$ to the adjacency $A^s_t$, and $\Lambda_t$ is the diagonal degree matrix with elements representing row summations of $A^s_t+I$. This normalization step ensures the adjacency to be positive semi-definite for spectral decomposition in GCN~\cite{kipf2016semi}. The ST-GCNN is further defined by convoluting the velocity map $V(l)$ with the kernel $W(l)$ at layer $l$ under the graph laplacian $\hat{A}^s_t$:
\begin{equation} 
f(V(l), A) = \sigma(\Lambda_t^{-\frac{1}{2}} \hat{A}^s_t \Lambda_t^{-\frac{1}{2}} V(l) W(l)).
\label{eq:stgcnnlayer}
\end{equation}

%The resulting embedding from the final ST-GCNN layer is denoted as $\overline{V}$ and is passed into the Time Extrapolator CNN. 

\subsection{The Time Extrapolator Convolution Neural Network (TXP-CNN)}
Following Social-STGCNN~\cite{mohamed2020social}, the goal of time extrapolator CNN is to conduct temporal convolutions on the past trajectory to decode the future trajectory, since temporal convolution network (TCN)~\cite{bai2018empirical} is considered as a more powerful and more efficient scheme to learn temporal dependencies than recurrent architecture. Specifically, we stack multiple convolutional layers along temporal domain of the output feature map $\hat{V}$ from the last ST-GCNN layer, with each temporal layer connected residually to its previous layer that is usually adopted to boost the network capacity~\cite{bai2018empirical}.

\subsection{Evaluation Metrics}
The two metrics used to evaluate the performance of the model are the Average Displacement Error (ADE) and the Final Displacement Error (FDE) introduced in \cite{pellegrini2009you}. ADE measures the performance of the model over all time steps in the prediction and computes the average error, while FDE only computes the error of the final time step. Since the proposed model outputs a bi-variate Gaussian distribution as in \cite{gupta2018social} and \cite{mohamed2020social}, we use the Minimum ADE (mADE) and Minimum FDE (mFDE) as in \cite{mohamed2020social} by taking $K$ samples from the predicted bi-variate Gaussian distribution with the minimum errors calculated as:
\begin{equation}
mADE_{K} = \frac{1}{N \times T_{pred}}\sum_{n=1}^{N} \min_{k}(\sum_{t=T_{obs}+1}^{T_{obs}+T_{pred}} ||\hat{p}^{nk}_t - p^n_t||_2),
\label{eq:made}
\end{equation}
\begin{equation}
mFDE_{K} = \frac{1}{N}\sum_{n=1}^{N}\min_{k}||\hat{p}^{nk}_{T_{obs}+T_{pred}} - p^n_{T_{obs}+T_{pred}}||_2,
\label{eq:mfde}
\end{equation}
where $\hat{p}^{nk}_t$ denotes the predicted position of the $k$th ($1\leq k\leq K$) sampled trajectory at time $t$, and $p^{n}$ is the ground truth trajectory. 

Since only picking the minimum error cannot show the robustness of the generated bi-variate Gaussian distributions, we introduce the new metrics known as the Average\textsuperscript{2} Displacement Error (aADE) and the Average Final Displacement Error (aFDE), which compare the models more holistically as below:
\begin{equation}
aADE = \frac{1}{N \times S \times T_{pred}}\sum_{n=1}^{N}\sum_{s=1}^{S} \sum_{t=T_{obs}+1}^{T_{obs}+T_{pred}} ||\hat{p}^{ns}_t - p^n_t||_2,
\label{eq:aade}
\end{equation}
\begin{equation}
aFDE = \frac{1}{N \times S}\sum_{n=1}^{N} \sum_{s=1}^{S} ||\hat{p}^{ns}_{T_{obs}+T_{pred}} - p^n_{T_{obs}+T_{pred}}||_2.
\label{eq:afde}
\end{equation} 
Note that since the ground truth trajectory distribution is unseen, we argue that aADE and aFDE are more accurate metrics than mADE and mFDE by taking $S$ samples of the predicted bi-variate Gaussian distribution into account and average the errors across these samples. 

\section{Experiments}
\subsection{Dataset}
We evaluate our multi-class model on the Stanford Drone dataset (SDD) \cite{robicquet2016learning} as it consists of both trajectories and class labels of different object classes. SDD %a large and widely used trajectory prediction dataset that 
contains 6 types of classes, i.e., \textit{biker}, \textit{pedestrian}, \textit{car}, \textit{cart}, \textit{bus}, and \textit{skater}. The scenes in SDD are shown in top views captured by drones. Following the existing works~\cite{gupta2018social,lee2017desire} tested on SDD, 8 frames are used as the past trajectory and the model will predict the next 12 frames, and 20 trajectories are randomly sampled ($K=20$) from the predicted multinomial distribution. We also adopt the metric mADE, mFDE, aADE, and aFDE, respectively, to test the effectiveness of our method.
%sampling rate

\subsection{Implementation Details}
Following Social-STGCNN~\cite{mohamed2020social}, PReLU is used as the activation function $\sigma(\cdot)$ in our model. The number of ST-GCNN and TXP-CNN layers are set to one and five respectively, with the best performance achieved. The training batch size is set to 512 with a learning rate of 0.0001 using the Adam optimizer. The input trajectory is normalized and denormalized with a scaling factor of 10. We also adopt the class balancing scheme to adjust the loss of different classes of objects according to their quantities. \textcolor{black}{The entire experiment is conducted on a NVIDIA 1080Ti GPU. The model is converged in 70 epochs and the training time of each epoch is around 4 minutes.}

\begin{figure*}
\centering
\begin{subfigure}{\textwidth}
  \centering
  \includegraphics[width=\linewidth]{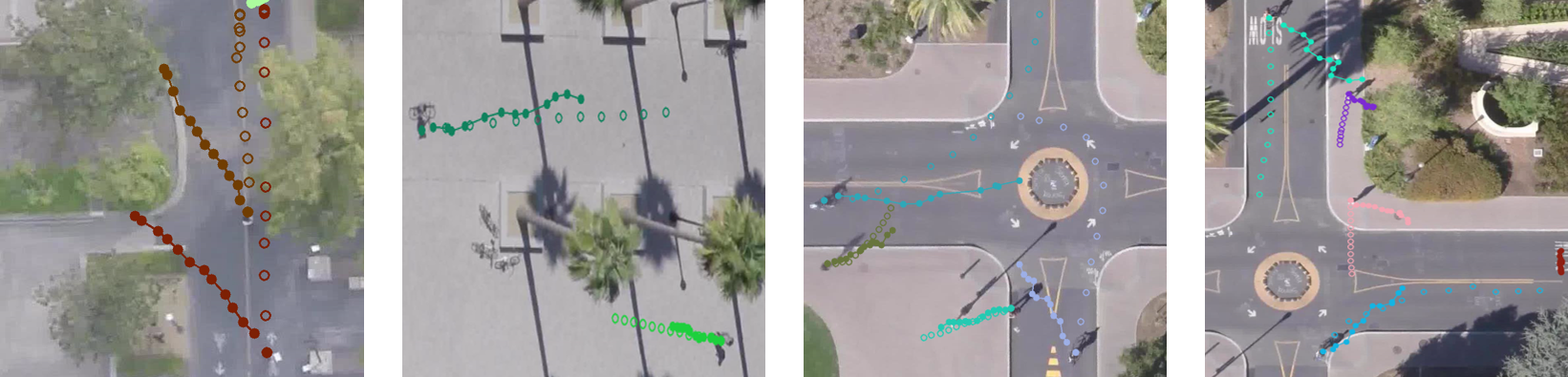}
  \caption{Social-STGCNN\vspace{2mm}}
\end{subfigure}%
\hfill
\begin{subfigure}{\textwidth}
  \centering
  \includegraphics[width=\linewidth]{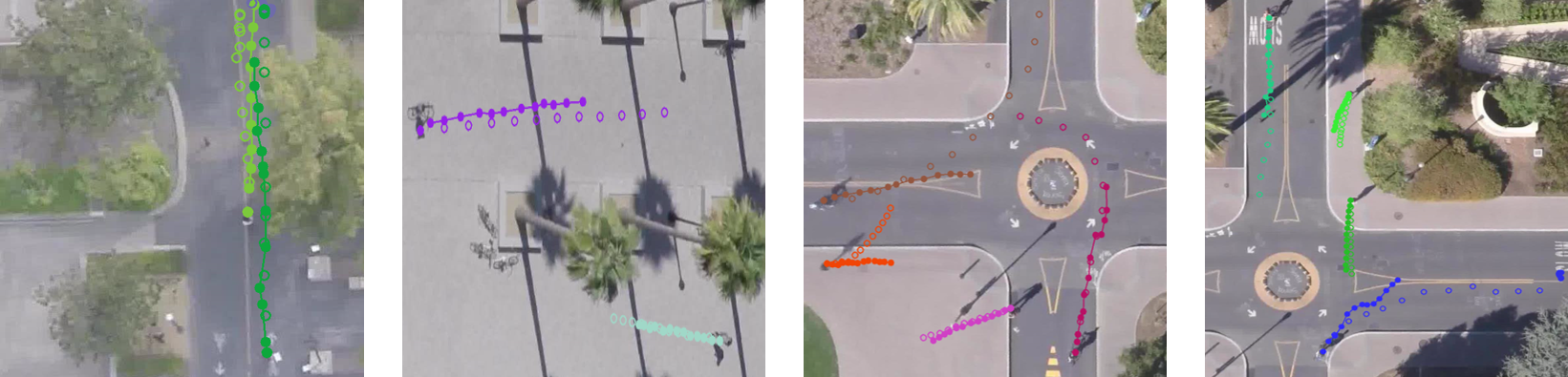}
  \caption{Semantics-STGCNN}
\end{subfigure}
 \caption{Qualitative comparisons of predicted trajectories in bird view of one frame of video by both Social STGCNN and Semantics-STGCNN. 3.2 seconds of past trajectory are used to predict 4.8 seconds of future trajectory, the ground truth can be seen in the hollow circle, and the line with filled circles is the prediction of the model.}
  \label{fig:vis_far}
\end{figure*}

\subsection{Quantitative Results}
\subsubsection{Compared with existing work} 
We first test our proposed Semantics-STGCNN that takes class information about a trajectory to help improve future trajectory accuracy and compare it to the existing methods. The compared baseline methods include a linear regressor (denoted as \textit{Linear}) to minimize the least square error, an energy minimization model (SF~\cite{yamaguchi2011you}) to regulate the path and social relationships, a recurrent model with pooling layer (Social-LSTM~\cite{alahi2016social}) to capture social dependencies, a GAN-based recurrent predictive model (Social-GAN~\cite{gupta2018social}) added on Social-LSTM, an attentive recurrent model (CAR-Net~\cite{sadeghian2018car}) that incorporates the information of saliency regions in the scenes, a conditional variational auto-encoder system (DESIRE~\cite{lee2017desire}) based on inverse optimal control (IOC), and Social-STGCNN~\cite{mohamed2020social}, which can be considered as a variant of our model without the multi-class label graph module. \textcolor{black}{The numerical results of Social-STGCNN are re-implemented on SDD by using their published code, and the other compared results are introduced from~\cite{lee2017desire} and~\cite{sadeghian2018car}.}

\begin{table}
\centering
\caption{Quantitative comparisons with mADE and mFDE in pixels.}
\begin{tabular}{ccc}
\hline
Method          & mADE           & mFDE           \\ \hline
Linear          & 37.11          & 63.51          \\ 
SF \cite{yamaguchi2011you}            & 36.48          & 58.14          \\ 
Social-LSTM \cite{alahi2016social}    & 31.19          & 56.97          \\ 
Social-GAN \cite{gupta2018social}      & 27.25          & 41.44          \\ 
CAR-Net \cite{sadeghian2018car}         & 25.72          & 51.80          \\ 
DESIRE \cite{lee2017desire}          & 19.25          & 34.05          \\ 
Social-STGCNN \cite{mohamed2020social}   & 26.46          & 42.71          \\ 
Semantics-STGCNN (Ours) & \textbf{18.12} & \textbf{29.70} \\ \hline
\end{tabular}
\label{tab:sota}
\end{table}

\begin{table}
\centering
\caption{Quantitative analysis with mADE, mFDE (minimum error) and aADE, and aFDE (average error).}% of our Semantics-STGCNN compared with Social-STGCNN.}
\resizebox{\columnwidth}{!}{%
\begin{tabular}{ccccc}
\hline
Method           & mADE           & mFDE           & aADE           & aFDE           \\ \hline
Social-STGCNN & 26.46          & 42.71          & 49.96          & 90.01          \\
Semantics-STGCNN (Ours) & \textbf{18.12} & \textbf{29.70} & \textbf{33.14} & \textbf{61.14} \\ \hline
\end{tabular}
}
\label{tab:minimum_average}
\end{table}

\begin{table*}
\centering
\caption{Quantitative analysis (averaged and per-class) with mADE and mFDE.}% for  of our Semantics-STGCNN compared with Social-STGCNN.}
\resizebox{\textwidth}{!}{%
\begin{tabular}{c||cc||cc|cc|cc|cc|cc|cc}
\hline
\multirow{2}{*}{Class} & \multicolumn{2}{c||}{Average} & \multicolumn{2}{c|}{Biker}       & \multicolumn{2}{c|}{Pedestrian} & \multicolumn{2}{c|}{Car}        & \multicolumn{2}{c|}{Bus}        & \multicolumn{2}{c|}{Skater}                           & \multicolumn{2}{c}{Cart}           \\
                       & mADE           & mFDE            & mADE           & mFDE           & mADE           & mFDE           & mADE           & mFDE           & \multicolumn{1}{c}{mADE}           & mFDE           & mADE           & mFDE           & mADE           & mFDE           \\ \hline
Social-STGCNN      & 40.98          & 66.18     & \textbf{54.86} & \textbf{100.96} & \textbf{21.13} & \textbf{32.39} & 32.68          & 51.12          & 59.76          & 84.21          & \multicolumn{1}{c}{48.62}          & 87.14          & \textbf{28.82} & \textbf{41.27}          \\ 
Semantics-STGCNN (Ours)             & \textbf{34.67} & \textbf{58.31}       & 69.38          & 113.17          & 42.25          & 72.29          & \textbf{13.24} & \textbf{20.49} & \textbf{18.31} & \textbf{35.78} & \multicolumn{1}{c}{\textbf{26.55}} & \textbf{40.11} & 38.27          & 67.99          \\ \hline
\end{tabular}%
}
\label{tab:per-class}
\end{table*}

The comparison performance is given in Table~\ref{tab:sota}. The results show that our Semantics-STGCNN performs the best with both the lowest average error (mADE) and the lowest final error (mFDE), which indicates that our predicted trajectory is more accurate at all time phases. We also observe that when comparing with Social-STGCNN, Semantics-STGCNN works better, especially in the long term with much lower mFDE. This shows that the class information is essential in improving the predicted path with fewer error accumulations.

\subsubsection{Minimum vs. Average}
We then show the prediction performance using our new average metrics of aADE (Eq.~\ref{eq:aade}) and aFDE (Eq.~\ref{eq:afde}) in Table~\ref{tab:minimum_average}. We first observe that our Semantics-STGCNN consistently outperforms Social-STGCNN in both minimum and average metrics. This shows that incorporating semantic information can effectively improve performance on average. We then observe the errors under average metrics (aADE and aFDE) are around two times larger than the minimum metrics (mADE and mFDE), respectively. This is because the minimum metrics only look at the best sample, which is less likely to happen in real-world scenarios, which makes it difficult to justify the model performance. While our new metrics, aADE, and aFDE, are more general to evaluate the prediction model by sampling over the generated path distributions. 

\subsubsection{Class-level Evaluations}
We also compare the error rates of Social-STGCNN and Semantics-STGCNN based on every class in Table~\ref{tab:per-class} to test how the label information will influence different types of trajectories. We first can see that Semantics-STGCNN achieves lower mADE and lower mFDE on average of all classes. The per-class results further illustrate that the items being faster-moving, i.e. \textit{Car}, \textit{Bus}, and \textit{Skater}, are more affected by the classes of the objects in the scene with a reasonable improvement in accuracy when class information is provided. This is because the objects with high speed can recognize the velocity dynamics from the ones within the same label, which will not confuse with the low-speed objects such as \textit{pedestrians} and \textit{bikers}.

\subsection{Qualitative Results}
\begin{figure*}
\centering
\begin{subfigure}{\textwidth}
  \centering
  \includegraphics[width=\linewidth]{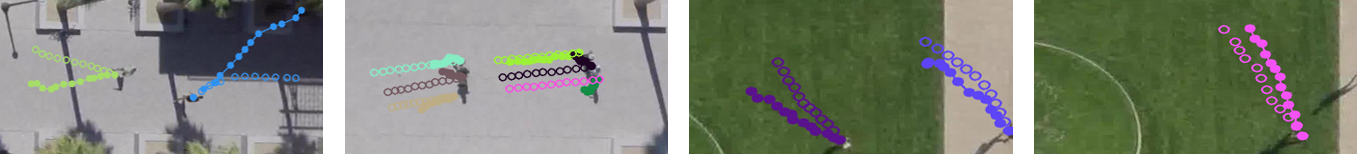}
  \caption{Social-STGCNN\vspace{2mm}}
\end{subfigure}%
\hfill
\begin{subfigure}{\textwidth}
  \centering
  \includegraphics[width=\linewidth]{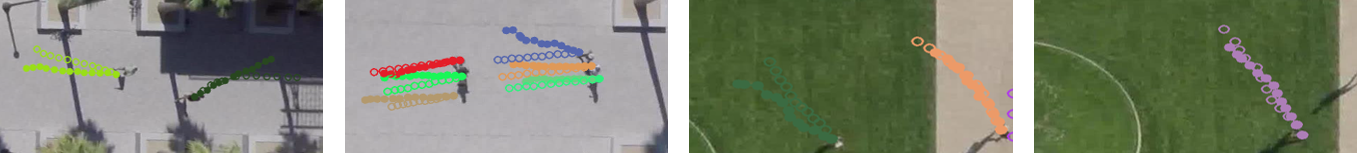}
  \caption{Semantics-STGCNN}
\end{subfigure}
 \caption{Qualitative comparisons of predicted trajectories in close view of one frame of video by both Social-STGCNN and Semantics-STGCNN.}
  \label{fig:vis_close}
\end{figure*}

Figure~\ref{fig:vis_far} compares the predictions of the two methods---Social-STGCNN and Semantics-STGCNN with example trajectories of \textit{bikers}, \textit{skaters}, and \textit{pedestrians}. In general, the visualized predictions of our model outperform Social-STGCNN, which shows consistent results with the quantitative evaluations. As in Fig.~\ref{fig:vis_far}, the prediction of Social-STGCNN drifts to a wrong direction, which results in a biased predicted path with large errors in the destination compared to the ground truth trajectory. For the samples of bikers, such as the purple path shown in the third subfigure of Fig.~\ref{fig:vis_far}(a), the distances of the consecutive predicted frames are rather close and inconsistent. This is because without observing the label of the trajectory, Social-STGCNN cannot effectively learn the differences among multiple types of trajectories, which causes inaccurate predictions with fast error accumulations in the long term. Other examples of Semantics-STGCNN outperforming Social STGCNN are demonstrated in Fig.~\ref{fig:vis_close}. In the result of Social-STGCNN from the second subfigure on the left-hand side, we observe that the predicted pedestrian trajectories are nearly stationary with high intensity at the starting points of the prediction. While for Semantics-STGCNN, we can clearly observe the moving predicted trajectories that show an active speed following the ground truth paths.

\begin{figure}
\centering
\begin{subfigure}{0.48\columnwidth}
  \centering
  \includegraphics[width=1\linewidth]{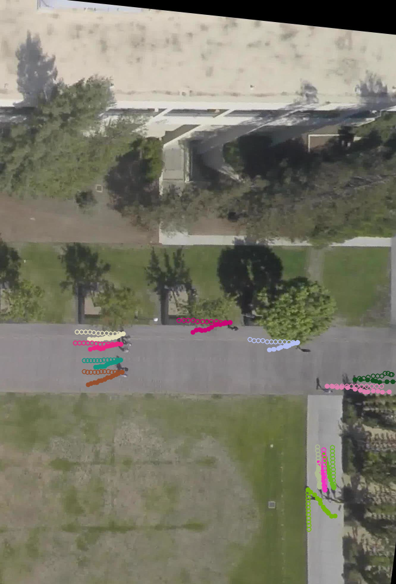}
  \caption{Social-STGCNN}
\end{subfigure}%
\hfill
\begin{subfigure}{0.48\columnwidth}
  \centering
  \includegraphics[width=1\linewidth]{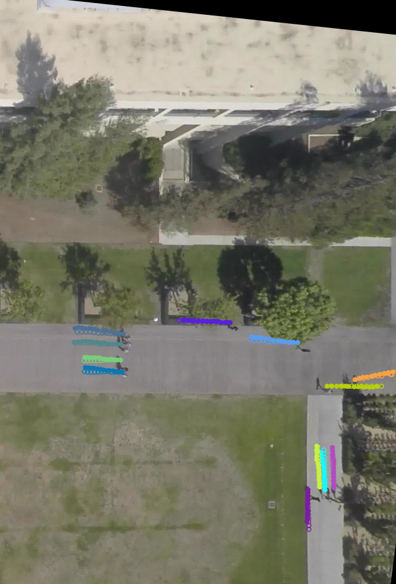}
  \caption{Semantics-STGCNN}
\end{subfigure}
 \caption{Qualitative comparisons of predicted trajectories in the whole scene (nexus) of one frame of video by both Social-STGCNN and Semantics-STGCNN.}
  \label{fig:vis_whole_1}
\end{figure}

\begin{figure}
\centering
\begin{subfigure}{0.48\columnwidth}
\centering
  \includegraphics[width=1\linewidth]{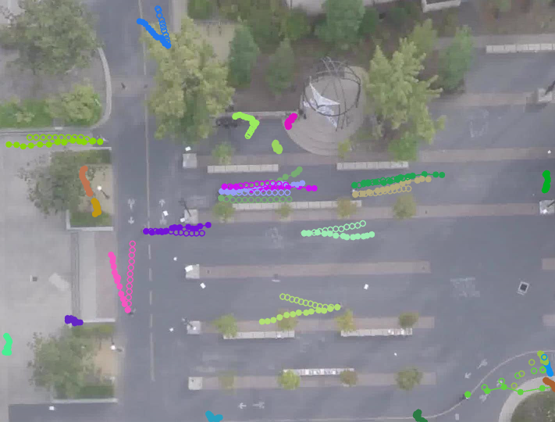}
  \caption{Social-STGCNN}
\end{subfigure}
\begin{subfigure}{0.48\columnwidth}
\centering
  \includegraphics[width=1\linewidth]{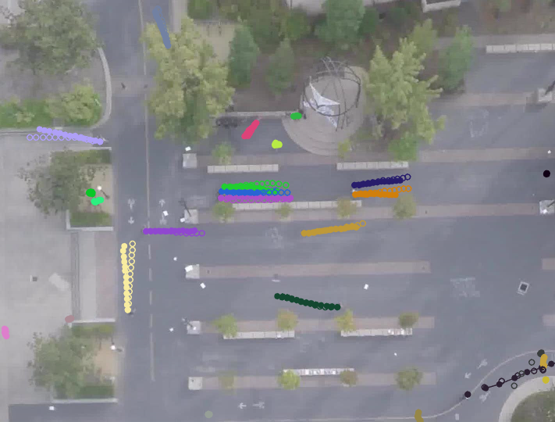}
  \caption{Semantics-STGCNN}
\end{subfigure}
 \caption{Qualitative comparisons of predicted trajectories in the whole scene (bookstore) of one frame of video by both Social-STGCNN and Semantics-STGCNN.}
  \label{fig:vis_whole_2}
\end{figure}

We further visualize the prediction results in the whole scene with more trajectories included. As compared in Fig.~\ref{fig:vis_whole_1}, Semantics-STGCNN ensures the prediction of more precise trajectories in high quality that better align with the real paths. Another example with diverse dynamics of the predictions is given in Fig.~\ref{fig:vis_whole_2}. From the pink and khaki dots in the left bottom of Fig.~\ref{fig:vis_whole_2}(b) comparing with the green and purple dots of Fig.~\ref{fig:vis_whole_2}(a), we observe that Semantics-STGCNN can better estimate the state of the static objects with fewer movements in the prediction. This comparison shows that our model effectively improves the performance of Social-STGCNN on both moving and static targets. Furthermore, the predicted paths of pedestrians for Social-STGCNN in the center of the scene are likely to collide with each other. In contrast, Semantics-STGCNN avoids this problem by introducing the class label in the graph adjacency, which ensures a more powerful social interaction by identifying the correlations of the existing paths of other pedestrians.

\subsection{Parameters \& Inference Speed}
By adding the class encoder module, we utilize fully-connected layers to the model to introduce the class label into the similarity matrix. This inevitably increases the model size and increases the inference time, but indeed the increase is not significant and will not impact real-time predictions nor storage capacity. Table \ref{tab:params} shows the number of qualities of both models. It shows an increase of 3.8\% in trainable parameters in our Semantics-STGCNN over Social-STGCNN, which only results in a slight increase of model size. We can also see that the inference time for both models are similar, and in fact, our model shows only a 2.3\% deficit compared to Social-STGCNN over a 20 sample average, despite adding more layers. This only increases our inference time on average by 7 milliseconds compared to Social-STGCNN. \textcolor{black}{Note that although the number of model parameters is positively correlated with the label size, the inference time is hardly affected. }%later is hardly increased in the real world. }

\begin{table}
\centering
\caption{Network comparisons with the number of trainable parameters, model size (bytes) and inference time (s).}% of Social-STGCNN and our Semantics-STGCNN.}
\resizebox{\columnwidth}{!}{%
\begin{tabular}{cccc}
\hline                                               Method &        Parameters   & Size (bytes) & Time (s)\\
                                      \hline
\multicolumn{1}{c}{Social-STGCNN}       & 7563                & 48,911                &     0.302     \\                     
\multicolumn{1}{c}{Semantics-STGCNN (Ours)} & 7852                & 43,631                &    0.309                          \\ \hline
\end{tabular}
}
\label{tab:params}
\end{table}

\section{Conclusion}
In this paper, we propose a Semantics-STGCNN framework to improve the performance of trajectory prediction under multiple classes by incorporating the semantic information from the object label. This is done by encoding an innovative semantics-guided adjacency by combining the label-based and velocity-based graph representations on the observed trajectory from different classes. We then integrate the spatial and temporal information by graph convolutions and temporal convolutions, respectively, to derive the Gaussian distributions of the future trajectory. The prediction results under various metrics show that by introducing label information in the graph, Semantics-STGCNN effectively improves the predicted trajectory with more accurate paths and fewer error accumulations and thus beats Social-STGCNN and the other state-of-the-art prediction models. \textcolor{black}{In the future, it would be beneficial by identifying the generally static objects (e.g., roads or trees) as the high-level label information to further aid the trajectory prediction. Furthermore, since our model is not limited to 2D video, it is also extendable to 3D trajectory with one more dimension included in the scene.}

\section*{Acknowledgement}
This work was supported in part by the Royal Society (Ref: IES\textbackslash R2\textbackslash 181024).


\begin{thebibliography}{10}
\providecommand{\url}[1]{#1}
\csname url@samestyle\endcsname
\providecommand{\newblock}{\relax}
\providecommand{\bibinfo}[2]{#2}
\providecommand{\BIBentrySTDinterwordspacing}{\spaceskip=0pt\relax}
\providecommand{\BIBentryALTinterwordstretchfactor}{4}
\providecommand{\BIBentryALTinterwordspacing}{\spaceskip=\fontdimen2\font plus
\BIBentryALTinterwordstretchfactor\fontdimen3\font minus
  \fontdimen4\font\relax}
\providecommand{\BIBforeignlanguage}[2]{{%
\expandafter\ifx\csname l@#1\endcsname\relax
\typeout{** WARNING: IEEEtran.bst: No hyphenation pattern has been}%
\typeout{** loaded for the language `#1'. Using the pattern for}%
\typeout{** the default language instead.}%
\else
\language=\csname l@#1\endcsname
\fi
#2}}
\providecommand{\BIBdecl}{\relax}
\BIBdecl



\bibitem{autocar}
A.~Houenou, P.~Bonnifait, V.~Cherfaoui, and W.~Yao, ``Vehicle Trajectory Prediction based on Motion Model and Maneuver Recognition,'' in \emph{IEEE International Conference on Intelligent Robots and Systems}, 2013, pp.
  4363--4369.

\bibitem{surveillance}
B.~Zhou, X.~Tang, and X.~Wang, ``Learning Collective Crowd Behaviors with
  Dynamic Pedestrian-Agents,'' \emph{International Journal of Computer Vision},
  vol. 111, no.~1, pp. 50--68, 2015.

\bibitem{mohamed2020social}
A.~Mohamed, K.~Qian, M.~Elhoseiny, and C.~Claudel, ``Social-STGCNN: A Social
  Spatio-Temporal Graph Convolutional Neural Network for Human Trajectory
  Prediction,'' in \emph{IEEE Conference on Computer Vision and Pattern
  Recognition}, 2020, pp. 14\,424--14\,432.

\bibitem{gupta2018social}
A.~Gupta, J.~Johnson, L.~Fei-Fei, S.~Savarese, and A.~Alahi, ``Social GAN:
  Socially Acceptable Trajectories with Generative Adversarial Networks,'' in
  \emph{IEEE Conference on Computer Vision and Pattern Recognition}, 2018, pp.
  2255--2264.

\bibitem{goodfellow2014generative}
I.~J. Goodfellow, J.~Pouget-Abadie, M.~Mirza, B.~Xu, D.~Warde-Farley, S.~Ozair,
  A.~Courville, and Y.~Bengio, ``Generative Adversarial Networks,'' \emph{arXiv
  preprint arXiv:1406.2661}, 2014.

\bibitem{yamaguchi2011you}
K.~Yamaguchi, A.~C. Berg, L.~E. Ortiz, and T.~L. Berg, ``Who Are You With and
  Where Are You Going?'' in \emph{IEEE Conference on Computer Vision and
  Pattern Recognition}, 2011, pp. 1345--1352.

\bibitem{alahi2016social}
A.~Alahi, K.~Goel, V.~Ramanathan, A.~Robicquet, L.~Fei-Fei, and S.~Savarese,
  ``Social LSTM: Human Trajectory Prediction in Crowded Spaces,'' in \emph{IEEE
  Conference on Computer Vision and Pattern Recognition}, 2016, pp. 961--971.

\bibitem{sadeghian2019sophie}
A.~Sadeghian, V.~Kosaraju, A.~Sadeghian, N.~Hirose, H.~Rezatofighi, and
  S.~Savarese, ``SoPhie: An Attentive GAN for Predicting Paths Compliant to Social and Physical Constraints,'' in \emph{IEEE Conference on Computer
  Vision and Pattern Recognition}, 2019, pp. 1349--1358.

\bibitem{rudenko2020human}
A.~Rudenko, L.~Palmieri, M.~Herman, K.~M. Kitani, D.~M. Gavrila, and K.~O.
  Arras, ``Human Motion Trajectory Prediction: A Survey,'' \emph{The
  International Journal of Robotics Research}, vol.~39, no.~8, pp. 895--935,
  2020.

\bibitem{robicquet2016learning}
A.~Robicquet, A.~Sadeghian, A.~Alahi, and S.~Savarese, ``Learning Social Etiquette: Human Trajectory Understanding in Crowded Scenes,'' in
  \emph{European Conference on Computer Vision}, 2016, pp. 549--565.

\bibitem{hochreiter1997long}
S.~Hochreiter and J.~Schmidhuber, ``Long Short-Term Memory,'' \emph{Neural
  Computation}, vol.~9, no.~8, pp. 1735--1780, 1997.

\bibitem{stgat}
Y.~Huang, H.~Bi, Z.~Li, T.~Mao, and Z.~Wang, ``STGAT: Modeling Spatial-Temporal Interactions for Human Trajectory Prediction,'' in \emph{IEEE International
  Conference on Computer Vision}, 2019, pp. 6272--6281.

\bibitem{kosaraju2019social}
V.~Kosaraju, A.~Sadeghian, R.~Mart{\'\i}n-Mart{\'\i}n, I.~D. Reid,
  H.~Rezatofighi, and S.~Savarese, ``Social-BiGAT: Multimodal Trajectory Forecasting using Bicycle-GAN and Graph Attention Networks,'' in
  \emph{Advances in Neural Information Processing Systems}, 2019.

\bibitem{kipf2016semi}
T.~N. Kipf and M.~Welling, ``Semi-Supervised Classification with Graph Convolutional Networks,'' \emph{arXiv preprint arXiv:1609.02907}, 2016.

\bibitem{lea2017temporal}
C.~Lea, M.~D. Flynn, R.~Vidal, A.~Reiter, and G.~D. Hager, ``Temporal
  Convolutional Networks for Action Segmentation and Detection,'' in \emph{IEEE
  Conference on Computer Vision and Pattern Recognition}, 2017, pp. 156--165.

\bibitem{longtermpred}
K.~Mangalam, Y.~An, H.~Girase, and J.~Malik, ``From Goals, Waypoints \& Paths To Long Term Human Trajectory Forecasting,'' \emph{arXiv preprint
  arXiv:2012.01526}, 2020.

\bibitem{pecnet}
K.~Mangalam, H.~Girase, S.~Agarwal, K.-H. Lee, E.~Adeli, J.~Malik, and
  A.~Gaidon, ``It Is Not the Journey but the Destination: Endpoint Conditioned Trajectory Prediction,'' in \emph{European Conference on Computer Vision},
  2020, pp. 759--776.

\bibitem{lee2017desire}
N.~Lee, W.~Choi, P.~Vernaza, C.~B. Choy, P.~H. Torr, and M.~Chandraker,
  ``DESIRE: Distant Future Prediction in Dynamic Scenes with Interacting Agents,'' in \emph{IEEE Conference on Computer Vision and Pattern
  Recognition}, 2017, pp. 336--345.

\bibitem{sadeghian2018car}
A.~Sadeghian, F.~Legros, M.~Voisin, R.~Vesel, A.~Alahi, and S.~Savarese,
  ``CAR-Net: Clairvoyant Attentive Recurrent Network,'' in \emph{European
  Conference on Computer Vision}, 2018, pp. 151--167.

\bibitem{zhang2020semantics}
P.~Zhang, C.~Lan, W.~Zeng, J.~Xing, J.~Xue, and N.~Zheng, ``Semantics-Guided Neural Networks for Efficient Skeleton-Based Human Action Recognition,'' in
  \emph{IEEE Conference on Computer Vision and Pattern Recognition}, 2020, pp.
  1112--1121.

\bibitem{bai2018empirical}
S.~Bai, J.~Z. Kolter, and V.~Koltun, ``An Empirical Evaluation of Generic Convolutional and Recurrent Networks for Sequence Modeling,'' \emph{arXiv
  preprint arXiv:1803.01271}, 2018.

\bibitem{pellegrini2009you}
S.~Pellegrini, A.~Ess, K.~Schindler, and L.~Van~Gool, ``You'll Never Walk Alone: Modeling Social Behavior for Multi-target Tracking,'' in \emph{IEEE
  International Conference on Computer Vision}, 2009, pp. 261--268.

\end{thebibliography}
\end{document}